%% file: main.tex
\documentclass{article}

\usepackage{microtype}
\usepackage{graphicx}
\usepackage{subfig}
\usepackage{amsmath}
\usepackage{amssymb}
\usepackage{booktabs} 
\usepackage[hyphens]{url}
\newif\ifarXiv
\arXivtrue 

\usepackage{hyperref}

\ifarXiv
    \usepackage[title]{appendix}
    \usepackage[accepted]{icml2020}
\else
    \usepackage{icml2020}
\fi

\icmltitlerunning{Classification Confidence Estimation with Test-Time Data-Augmentation}

\DeclareMathOperator*{\argmax}{argmax}

\author{Yuval Bahat and Gregory Shakhnarovich\\
Technion - Israel Institute of Technology, Haifa, Israel\\
{\tt\small \{yuval.bahat@campus,tomer.m@ee\}.technion.ac.il}}

\begin{document}

\twocolumn[
\icmltitle{Classification Confidence Estimation with Test-Time Data-Augmentation}
\begin{icmlauthorlist}
\icmlauthor{Yuval Bahat}{technion}
\icmlauthor{Gregory Shakhnarovich}{TTIC}
\end{icmlauthorlist}

\icmlaffiliation{technion}{Technion - Israel Institute of Technology, Haifa, Israel}
\icmlaffiliation{TTIC}{Toyoyta Technological Institute at Chicago, Chicago, IL, USA}

\icmlcorrespondingauthor{Yuval Bahat}{yuval.bahat@gmail.com}
\icmlcorrespondingauthor{Gregory Shakhnarovich}{gregory@ttic.edu}




\vskip 0.3in
]


\printAffiliationsAndNotice{}  

\input{abstract.tex}
\input{intro.tex}
\input{related_work.tex}
\input{method.tex}

\input{experiments.tex}
\input{conclusion.tex}

\ifarXiv
    \onecolumn
    \appendixpage
    \input{supplementary_sections}
\fi

\bibliography{main}
\bibliographystyle{icml2020}

\end{document}

%% file: abstract.tex
\begin{abstract}
    Machine learning plays an increasingly significant role in many aspects of our lives
    (including medicine, transportation, security, justice and other domains), making the potential consequences of false predictions increasingly devastating. These consequences may be mitigated if we can automatically flag such false predictions and potentially assign them to alternative, more reliable mechanisms, that are possibly more costly and involve human attention. 
    This suggests the task of detecting errors, which we tackle in this paper for the case of visual classification.
    To this end, we propose a novel approach for classification confidence estimation. We apply a set of semantics-preserving image transformations to the input image, and show how the resulting image sets can be used to estimate confidence in the classifier's prediction. 
    We demonstrate the potential of our approach by extensively evaluating it on a wide variety of classifier architectures and datasets, including ResNext/ImageNet, achieving state of the art performance.
    This paper constitutes a significant revision of our earlier work in this direction \cite{bahat2018invariance}.
\end{abstract}{}

%% file: intro.tex
\section{Introduction}
\input{fig_synthesizing_images.tex}
Despite rapid and continuing dramatic improvements in the accuracy of predictors applied to computer vision, these predictors (image classifiers, object detectors, etc.) continue to have non-negligible error rates. Prediction errors can have a critical effect in sensitive applications (e.g. medical, autonomous driving, security). Thus, developing means for preventing prediction errors has drawn substantial research attention.

One may aim to mitigate the problem by improving predictors' accuracy. 
This paper relates to the complementary research direction of estimating prediction confidence. Our focus is on means to assess the degree of prediction uncertainty, at inference (test) time, and for a given test instance. This, in turn, can be used to endow predictors with a  rejection ability. When estimated uncertainty exceeds a certain threshold, an instance may be flagged as one the predictor is unable to classify. 

This ability is desirable for many applications. For instance when using a medical classifier to identify malignant cell tissues, a reliable classification uncertainty estimate would allow sending poorly classified tissues for a thorough human examination, thus avoiding the grave consequences of a misclassification. 
Our experiments in this paper involve image classifiers using convolutional networks,  although our ideas can be easily extended to other tasks and modalities.

Classification errors can be generally divided into three different types. As image classifiers are trained on a pre-defined set of image classes, images whose correct class falls outside this set ("out of distribution") are bound to be misclassified. A second error type concerns images misclassified despite belonging to a class in the classifier's training set. The third type comprises images generated by an adversary in a deliberate effort to ``fool'' the classifier \cite{szegedy2013intriguing,nguyen2015easily_fooled}. Our work focuses on the first two (non-adversarial) error types.

The key idea behind our approach is to estimate the confidence of a classifier's prediction on a given image by assessing the classifier's \emph{scene-specific accuracy}.
The standard notion of classifier error captures the expected accuracy on the entire distribution of images. It is normally estimated by empirical accuracy over a validation or test set. 
In contrast, the scene-specific accuracy we consider, quantifies the classifier's performance in the limited context of images depicting the specific object or scene appearing in the given image.

To estimate such scene-specific accuracy we address two challenges: (i) with only one given image, we need to "manufacture" additional images depicting the same content, and (ii) at test time, we do not have the ground truth label for that image; so how can we estimate the accuracy?

To address these challenges, we demonstrate how certain image transformations (e.g. horizontal flip, or shift), commonly used as training data augmentation techniques, can be applied to the input image to form a set of synthesized images, serving as a proxy for the anticipated scene-conditional image distribution (see Fig.~\ref{fig:synthesizing_images}). We then propose a technique to aggregate classifier's outputs corresponding to these synthetic images into a classification confidence estimate (without access to ground-truth labels). 
A byproduct of this framework is a potential gain to the classifier's classification accuracy, by considering the entire set of transformed images in its prediction process.

Our proposed confidence estimation framework requires no training, and incurs only a small computational overhead. It can be applied to any given input image and any "off the shelf" image classifier. No access to the internal operation of the classifier is required.
Our experiments show this framework is beneficial for yielding state of the art (SOTA) error detection performance, as we demonstrate on multiple image datasets using SOTA classifier architectures.

To summarize, the contributions in this work include: (i) a novel confidence estimation method based on applying image transformations for test-time data augmentation. This method is applicable to any pre-trained image classifier without requiring modifications or access to its internal layers. 
(ii) A thorough evaluation of our method on a variety of datasets and classifiers, demonstrating superior confidence estimation performance compared to the current SOTA. 
(iii) A further improvement in confidence estimation performance by using a bootstrapping technique, combined with a novel ranking algorithm.
(iv) A modification to the existing confidence estimation performance metric that makes it more informative and interpretable.

%% file: fig_synthesizing_images.tex
\begin{figure*}[t]
\centering
\includegraphics[width=\textwidth]{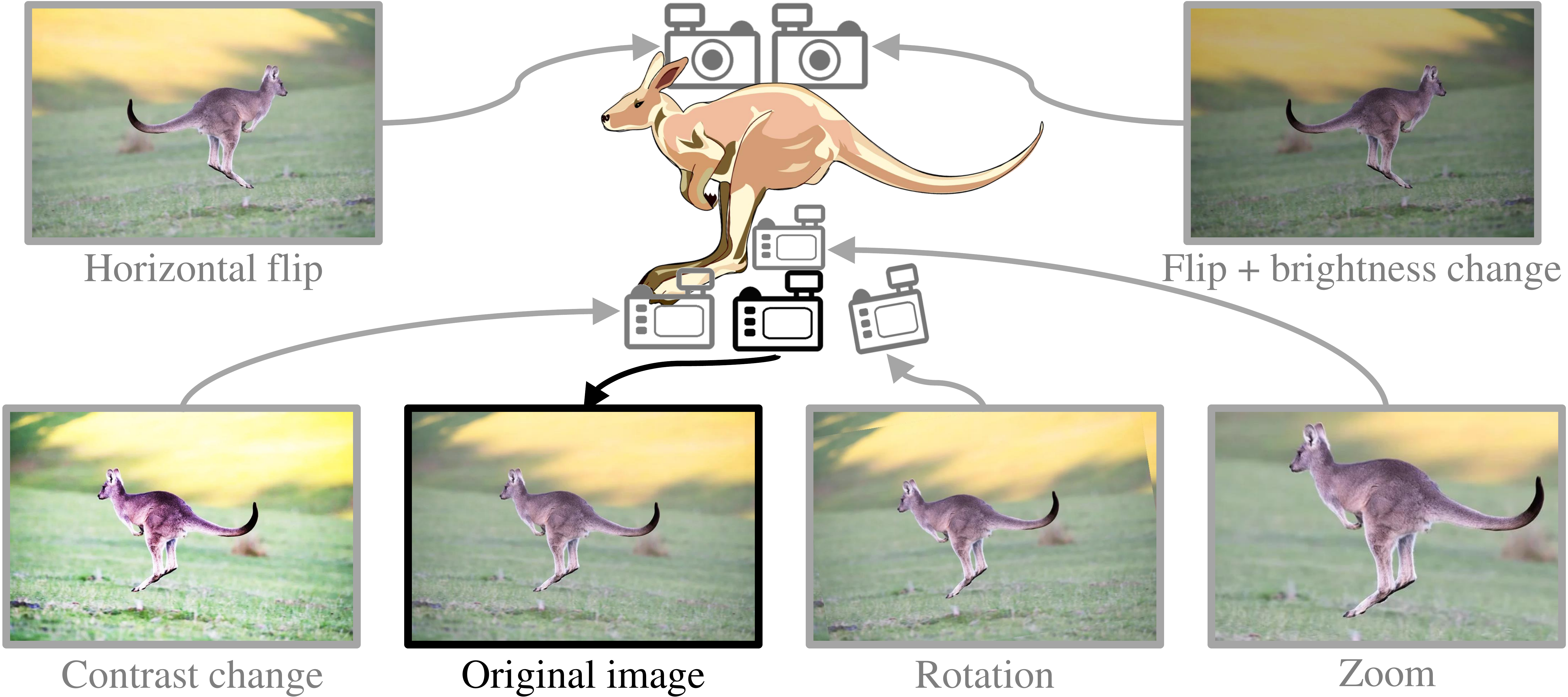}
\vspace{-20pt}
\caption{
\textbf{Using image transformations to simulate different imaging conditions.}
We apply a set of semantic preserving image transformations on the input image, and then use the resulting image set to estimate the classifier's reliability in the limited context of the specific given image content.}
\vspace{-10pt}
\label{fig:synthesizing_images}
\end{figure*}

%% file: related_work.tex
\section{Background and Related Work}
Given an input image $x$, the output of a classifier $f$ trained to recognize $C$ different classes is a vector $s(x)=f(x)$ whose pseudo-probability values $\{s_c(x)\}_{c=1}^{C}$ result from a Softmax operator applied on its last layer. The predicted class is then set to be \mbox{$\hat{c}(x)=\argmax_cs_c(x)$}. A classification confidence predictor then attempts to estimate the probability of the predicted class $\hat{c}(x)$ matching the correct class $c(x)$.

Classification confidence estimation can be broadly divided into two use cases, namely \emph{confidence calibration} and \emph{selective classification} \cite{ding2019uncertainty_two_cases}. Methods related to the first use case focus on adjusting the classifier's outputs corresponding to each possible class, to better reflect the true posterior probability. Our approach falls under the second use case, related to the reject option, where thresholding the estimated confidence allows for discarding under-confident classifications.

The \emph{Maximal Softmax Response} (MSR) method, suggested by \cite{cordella1995pre_msr} and recently revisited by \cite{hendrycks2016msr}, uses the classifier's maximal softmax output $\max_cs_c(x)$ (corresponding to the predicted class) as the classification confidence score. Despite softmax values being uncalibrated, this baseline method performs very well in the context of selective classification. Incorporating the use of the maximal softmax response into our novel approach yields superior confidence estimation performance.

Bayesian neural networks (BNNs) \cite{mackay1992bayesian_neural_net} constitute a more sophisticated approach for confidence estimation, where neural network (NN) parameters are represented as probability distributions, allowing them to provide a confidence interval around their non-deterministic outputs by repeating each classification multiple times. However, this approach has trouble scaling to the size of practical datasets and NNs. Hern{\'a}ndez-Lobato \& Adams \yrcite{hernandez2015scalable_bnn} proposed a more scalable training algorithm and Lakshminarayanan et al. \yrcite{lakshminarayanan2017deep_ensembles} proposed to approximate BNNs using an ensemble of NNs (NN-Ens), randomly initializing the weights of each NN using a different seed. The former method was only evaluated on the confidence calibration use case, and more importantly, both methods require training their NN classifiers from scratch. In contrast, our method utilizes an ``ensemble'' of simple image transformations instead, and can therefore be applied to any given classifier, not requiring any further training. Nonetheless, we show in our experiments the benefit of using our proposed method even when 
utilizing the NN-Ens method. 

An even better scaling proxy to BNNs is the Monte-Carlo dropout (MCD) method \cite{gal2016mc_dropout}, that suggests performing multiple repetitions of each image classification at test time, each time randomly dropping other activation units. This method requires no additional training but incurs some computational overhead, and
requires access to the NN's internal layers. Geifman \& El-Yaniv \yrcite{geifman2017selective_classification} evaluated its effectiveness for confidence estimation by regarding the variance of the classifier's output $s_{\hat{c}(x)}(x)$ across different repetitions as an uncertainty score\footnote{The opposite of the confidence score.}, and found it to be on a par with the MSR method on the CIFAR datasets and significantly below MSR on ImageNet.

Other proposed methods either take a more invasive approach and use internal activation distances \cite{mandelbaum2017distance_confidence} (IAD), or use hand-crafted features \cite{Song_2018attributes_confidence,brunner2019mandelbaum_like_embeddings} to infer confidence. A recent comparison \cite{geifman2018biasreduced} of the IAD, MSR, MCD and NN-Ens methods on the CIFAR, SVHN and ImageNet datasets showed the NN-Ens method performed best on all datasets, closely followed by the MSR baseline. Among these four methods, MSR is the only method that can be applied to any given classifier without further training or having access to its internal layers; so can our approach.

%% file: method.tex
\section{Proposed Method}
Recall that the accuracy of an image classifier $f$ is
\begin{equation}
\label{eq:F_acc}
\text{Acc}_f=\frac{1}{|D_f|}\sum_{x\in D_f}{\cal I}[\hat{c}(x),c(x)],
\end{equation}
where $D_f$ is a set of images approximating the distribution of images in $f$'s intended domain (e.g. the ImageNet validation set), and ${\cal I}[\hat{c}(x),c(x)]$ is an indicator function returning $1$ for correct classifications ($\hat{c}(x)==c(x)$) and $0$ otherwise. 
\subsection{Estimating Classification Confidence}\label{sec:method_conf_est}
Unlike the standard accuracy measure, in the context of estimating a classification confidence score
we are interested in the accuracy of $f$ when operating on the \emph{specific content depicted in $x$}. We denote this content (e.g. an object or scene) by $\chi$, and propose to estimate the classification confidence by viewing $x$ as a random sample from a distribution of images capturing $\chi$ under different plausible imaging conditions, including different capturing angle, different illumination settings, etc. Having access to a set of images $D_\chi$ approximating this distribution can be used to estimate $f$'s accuracy when operating on images depicting $\chi$, similarly to the way $D_f$ is used when calculating the standard classifier accuracy on its entire domain.

This approach, however, poses two principle challenges, namely (i) obtaining more images of $\chi$ in addition to $x$ and (ii) estimating classification correctness for each of the images in $D_\chi$, to substitute for indicator function ${\cal I}[\hat{c}(x),c(x)]$ in \eqref{eq:F_acc}, as the ground truth label corresponding to $x$ is obviously unavailable.
To tackle the first challenge, we synthesize new images by applying a set of semantics-preserving image transformations $\cal T$ on $x$, yielding image set $D_\chi=\{x_i\}_{i=0}^{|{\cal T}|}$, where $x_i$ is the result of applying transformation ${\cal T}\{i\}$ on $x$ and ${\cal T}\{0\}$ is set to the identity transformation ($x_0=x$).

We use these transformations, commonly used as data augmentations techniques for NN training \cite{krizhevsky2012data_augm}, to simulate capturing $\chi$ under different imaging conditions. For example, applying horizontal image flipping or image zoom-in by a factor $\alpha$ simulate capturing $\chi$ from the opposite direction or using different camera zoom, respectively, as illustrated in Fig.~\ref{fig:synthesizing_images}. Other transformations in $\cal T$ include various image translations, contrast manipulations and Gamma corrections. We avoid transformations that correspond to unlikely imaging conditions, such as vertical image flipping. Note that the employed transformations should be adapted to the domain of $f$, e.g. by avoiding horizontal image flipping in case $f$ is a numeric digit classifier; we will revisit this idea in Sec.~\ref{sec:exp_transformations}.

Next, we apply classifier $f$ on all images in $D_\chi$ and compute its averaged softmax output \mbox{$s(D_\chi)=\frac{1}{|D_\chi|}\sum_{x_i\in D_\chi}s(x_i)$}. Then we take the softmax value corresponding to the chosen class $\hat{c}(x)=\argmax_c\{s_c(x)\}$ to be our confidence estimate:
\begin{equation}\label{eq:ens_MSR_conf}
    \hat{r}_x^f=s_{\hat{c}(x)}(D_\chi)=\frac{1}{|D_\chi|}\sum_{x_i\in D_\chi}s_{\hat{c}(x)}(x_i).
\end{equation}{}

We note that the transformed image versions in $D_\chi$ can additionally be used for improving the classification accuracy, by choosing the predicted class to be \mbox{$\hat{c}_\chi(x) = \argmax_c\sum_{x_i\in D_\chi}s_c(x)$}.
This is known as test-time data augmentation, and was originally proposed by \cite{simonyan2014test_time_augm}.
Since using this scheme may result in different (potentially more accurate) classifications compared to those based solely on $x$, 
our confidence estimate in \eqref{eq:ens_MSR_conf} should be adapted accordingly, by replacing the index $\hat{c}(x)$ with $\hat{c}_\chi(x)$. 

\paragraph{Bootstrapping for enhanced confidence estimation.}
Applying image transformations on $x$ is a surrogate for sampling from a distribution of possible images depicting content $\chi$. However, this surrogate depends on the set of chosen transformations $\cal T$, and therefore does not necessarily reflect the actual, unknown distribution of possible images. To overcome this problem and improve the reliability of our classification confidence estimate, we propose to adopt a bootstrapping resampling technique \cite{efron1992bootstrap} and compose $N_{BS}$ different sets of $|D_\chi|$ images, each sampled at random with replacements from $D_\chi$. Applying our proposed confidence estimator \eqref{eq:ens_MSR_conf} on each of these sets yields a set of confidence scores $\{\hat{r}_{x,b}^f\}_{b=1}^{N_{BS}}$ corresponding to the classification $f(x)$. This can then be used, e.g., for providing a confidence interval around our classification confidence score. We can also harness this set in the context of selective classification, as we explain in Sec.~\ref{sec:selective_classification}. Note that using bootstrapping implicitly assumes samples are independent, while this is not the case with images in $D_\chi$, that were all modified from a single image $x$. Nonetheless, using this technique can improve our confidence estimation performance (see Sec.~\ref{sec:experiments}).

\subsection{Confidence Estimation for Selective Classification}\label{sec:selective_classification}
We evaluate our confidence estimation method in the context of selective prediction \cite{elyaniv2010selective_classification1,wiener2011selective_classification2}, which we briefly explain next. Given a set of image classifications and corresponding classification confidence scores, a selective classification mechanism indicates which classifications should be rejected, based on a given minimal confidence threshold or desired classification coverage. The latter case (involving a desired coverage parameter) was recently used for comparing the performance of different classification confidence estimation methods \cite{ding2019uncertainty_two_cases,geifman2018biasreduced}, by measuring the resulting error rate (risk) over the non-rejected images. A faultless confidence estimator would assign higher scores to all correct classification, resulting in a zero risk as long as the desired coverage is lower than the given classifier's accuracy. In contrast, the worst possible confidence estimator would make an opposite confidence scores assignment, causing the selection mechanism to favor incorrect classifications over correct ones. 
This suggests that the performance of a classification confidence method measured in the context of selective classification only depends on the method's ability to rank classifications in the correct confidence order, regardless of the absolute assigned confidence scores.

\paragraph{Adapting the bootstrapped scores for selective classification.}
To exploit our proposed bootstrapping scheme in the selective classification use case, we need to convert the $N_{BS}$ confidence scores corresponding to each image classification into a reliable ranking of the classifications' confidence. To this end, we next propose a novel \emph{sliding window plurality} algorithm:
\begin{enumerate}
    \vspace{-10pt}
    \item Let $N$ be the number of images, where bootstrapping yields $N_{BS}$ confidence scores per image. We sort all $N\cdot N_{BS}$ scores in decreasing order into a list, where we store the index of the image associated with each score. We expect the indices corresponding to reliably classified images to appear earlier in the list than those corresponding to incorrectly classified images.
    \vspace{-7pt}
    \item\label{enum:loop_start} We observe the first $W_{BS}$ indices in the list, and look for the one with the highest number of occurrences within these $W_{BS}$ indices. 
    \vspace{-7pt}
    \item The image classification corresponding to this index is assigned the top confidence rank.
    \vspace{-7pt}
    \item We remove all occurrences of this index throughout the list.
    \vspace{-7pt}
    \item\label{enum:loop_end} We go back to step~\ref{enum:loop_start}, looking for the next most frequent index at the top of the list.
    \vspace{-7pt}
    \item We iterate through steps~\ref{enum:loop_start}-\ref{enum:loop_end}, each time assigning the current most frequent index the next confidence rank. 
\end{enumerate}    \vspace{-7pt}

%% file: experiments.tex
\section{Experiments}\label{sec:experiments}
To evaluate the proposed confidence estimation method, we focused on the context of selective-classification and tested our method's performance on a wide variety of datasets and classification architectures. Our evaluation datasets range from the CIFAR-10 \& CIFAR-100 datasets of tiny ($32\times32$ pixels) natural images  \cite{krizhevsky2009cifar}, through the SVHN dataset \cite{netzer2011svhn} depicting house number digits, to the realistic images of the STL-10 \cite{coates2011stl10} and ImageNet \cite{russakovsky2015imagenet} datasets. We employed a wide variety of classifier architectures, ranging from the top performing ones to those exhibiting lower accuracy, like the ResNext \cite{xie2016resnext} and AlexNet \cite{krizhevsky2012data_augm} architectures, respectively (in the case of ImageNet). 
\input{fig_AORC_demo.tex}
Our method demonstrated superior confidence estimation performance throughout these tests, as we report next.

\subsection{Evaluation metrics}\label{sec:eval_metrics}
Earlier works (e.g. \cite{mandelbaum2017distance_confidence}) used the \emph{area under Receiver Operator Characteristics} (AUROC) and the \emph{area under Precision Recall} (AUPR) metrics for comparing confidence estimation performance. In this paper we follow the recent convention laid by \cite{el2010risk_coverage} and adopted by \cite{ding2019uncertainty_two_cases}, who showed that AUROC or AUPR scores cannot be compared unless evaluated on the exact same classifier. They instead proposed to use the \emph{area under risk coverage} (AURC) curve measure, directly related to the selective classification application discussed in Sec.~\ref{sec:selective_classification}. Each point on the risk-coverage curve corresponds to a certain confidence threshold, inducing truthfully and falsely detected classification errors (positive detections), denoted TP and FP respectively, and their complementary terms for truthfully or falsely identified correct classifications (negative detections), TN and FN. A (coverage,risk) point coordinates are then calculated as:
\begin{equation}
    \label{eq:RC_definitions}
    \text{\hspace{-.8em}coverage}=\frac{\text{TN}+\text{FN}}{\text{TN}+\text{FN}+\text{TP}+\text{FP}},
    ~\text{risk}=\frac{\text{FN}}{\text{TN}+\text{FN}}.
\end{equation}{}

\vspace{-10pt}
The risk-coverage curve is then plotted by recording the cumulative error rate observed when sequentially classifying images according to a decreasing classification confidence ranking. Each (coverage,risk) point on this curve therefore corresponds to the error rate (risk) incurred when accepting only the portion of the top ranked images corresponding to the coverage value, and rejecting the rest.

Geifman et al. \yrcite{geifman2018biasreduced} have recently proposed to normalize AURC by subtracting the inevitable risk incurred when using a faultless (oracle) confidence estimator, denoting it \emph{excess-AURC}. We take another step to make evaluation scores more informative, by measuring the area lying \emph{over} the excess coverage-risk (RC) curve and under an RC curve corresponding to ranking according to the worst possible confidence estimator, and dividing it by the entire area between the worst and faultless estimator curves, as illustrated in Fig.~\ref{fig:AORC_demo}.
This proposed metric (denoted AORC) directly (rather than inversely) corresponds to confidence estimation performance, making it arguably more intuitive. More importantly, being adjusted to the worst possible estimator brings the AORC score to lie in $[0,1]$ regardless of the accuracy of the used classifier, in contrast to excess-AURC.

\subsection{Experimental Setup}\label{sec:exp_setup}
Throughout our experiments, we first classify the images of a certain dataset using a pre-trained image classifier, and then apply the evaluated confidence estimation method on the resulting classifications. 
We emphasize that we are only interested in the performance of classification confidence estimation, and not in the performance of the classifiers, whose accuracy we take as a given.
We used pre-trained  VGG-8 classifiers~\cite{vgg-online} for the CIFAR-10, CIFAR-100 and SVHN dataset, 
and the pre-trained AlexNet, ResNet18 and ResNext classifiers~\cite{pytorch_torchvision_models}
 for ImageNet.
For STL-10, we used an ELU classifier \cite{clevert2015elu_model} trained by \cite{mandelbaum2017distance_confidence}, to allow a comparison with their method, and additionally trained several state-of-the-art Wide-ResNet models \cite{zagoruyko2016wide_resnet}, to allow further examining specific scenarios.
Evaluation is performed on the validation partitions of each of the datasets, where we first separate a small portion of the dataset to allow experimenting with different transformation sets $\cal T$  in Sec.~\ref{sec:exp_transformations}. We refer to this subset as the experimental subset, constituting 2\%-10\% of the images in the different datasets, while we use the rest of the validation set images as our evaluation subset, in Sec.~\ref{sec:exp_performance}.\footnote{We will make the exact partition of each validation set and our pre-trained classifiers available, along with our code.}
\input{fig_per_param_AORC.tex}

\begin{table}[t!]
    \centering
    \input{tab_transformations_original_post.tex}
    \caption{\textbf{Comparing image transformations.} AORC scores correspond to confidence estimation by taking the MSR of a classifier's output corresponding to a single transformed image. Presented AORC values are multiplied by $100$ for clarity. Transformations, per row: Identity (original $x$), horizontal flip, downwards $5$ pixels shift, upwards $5$ pixels shift, hor. flip followed by upward shift, $3^\circ$ rotation, vertical flip and color channel swapping. The right-most two columns correspond to STL-10 classifiers trained with vs. without hor. flip data augmentation.
    As one might expect, the notion of `natural' transformations is dataset-specific, as demonstrated by the low AORC score achieved when employing the horizontal flipping transformation on the SVHN digits dataset or the color-channels swapping transformation on the STL-10 dataset of natural images.}
    \label{tab:different_transformations}
\end{table}{}

\subsection{Image Transformations}\label{sec:exp_transformations}
Applying a set of natural image transformations $\cal T$ on $x$ should yield a set of images $D_\chi$ that are likely to be acquired when capturing content $\chi$ under natural circumstances. However, the notion of \emph{natural} transformations varies across domains, and therefore the chosen $\cal T$ can be dataset-specific.
Tab.~\ref{tab:different_transformations} presents AORC scores corresponding to confidence estimation computed using a single transformation $t$, by taking the MSR corresponding to $t(x)$. For some transformations, performance is consistent across datasets, like the high-scoring shift transformation (with different directions) and the low-scoring, unnatural vertical flip transformation. For other transformations however, scores vary across datasets, like the horizontal image flip, that yields good performance on CIFAR and STL but not on the SVHN digits dataset, for which horizontal flipping is obviously not natural. Similarly, color channel swapping significantly degrades performance for the natural image datasets, but not for SVHN. The two right-most columns in Tab.~\ref{tab:different_transformations} compare STL-10 Wide ResNet classifier trained with vs. without horizontal flipping data augmentation, demonstrating a relatively small effect on estimation performance, even when utilizing the horizontal flipping transformation itself at test time (2\textsuperscript{nd} and 5\textsuperscript{th} rows).

Some transformation types constitute an infinite set of possible transformations, that can be characterized using one or more parameters. The effect of these parameter on confidence estimation performance is depicted in Fig.~\ref{fig:per_param_AORC} for the image shift and Gamma correction\footnote{Gamma correction is the action of raising each image pixel to the power of $\gamma$, where image pixels are in $[0,1]$.} transformation cases. We followed the same procedure conducted for Tab.~\ref{tab:different_transformations} and computed per-parameter AORC values, each corresponding to confidence estimates utilizing a single image transformation. Similarly to color channel swapping, Gamma transformation seems to be useful for confidence estimation in the SVHN case, but less so for natural images (in which case AORC values peak around $\gamma=1$, corresponding to the original image $x$). AORC values corresponding to a certain relative shift radius (bottom figure) were obtained by averaging over shifting in multiple directions. For the CIFAR-100 and STL-10 dataset curves, we average  over shifting preceded by horizontal flipping too. The plots suggest that confidence estimation performance and shift magnitude are inversely related, across the examined datasets.

The experiments described thus far involved only a single transformed image, to isolate the effect of different transformations $t$.
To achieve a significant performance gain, we need to consider the larger set $D_\chi$ created using a set of transformations $\cal T$. As suggested by our experiments so far, the optimal $\cal T$ can be different for each dataset, and the task of finding this optimal set can be non-trivial.
In this work we used the experimental subsets (see Sec.~\ref{sec:exp_setup}) to choose a specific set $\cal T$ per dataset (across its different classifiers), chosen to yield good confidence estimation performance. We report specific details on the exact set $\cal T$ and bootstrapping parameters $N_{BS}$, $W_{BS}$ used for each dataset in the supplementary material. An interesting goal for future work is optimizing and automating this choice for a given dataset.

\subsection{Performance Comparison}\label{sec:exp_performance}
We compared our method, with and without using bootstrapping, with the original MSR method by Hendrycks \& Gimpel \yrcite{hendrycks2016msr}, as both methods consider the classifier to be a `black-box' and require no access or modifications to its internal components. We refrain from comparing to the MC-dropout method \cite{gal2016mc_dropout}, as it requires modifying the classifiers' mode of operation, and because it was found in recent works by Geifman et al. \yrcite{geifman2017selective_classification,geifman2018biasreduced} to be outperformed by MSR.

We use the suggested AORC score to report results in all the following experiments, which correspond to the case of an image classifier predicting according to $x$ alone, rather than based on all $x_i\in D_\chi$ (see Sec.~\ref{sec:method_conf_est}). 
Nonetheless, we emphasize that the advantage of our method remains unchanged when the classifier relies on the entire set $D_\chi$, or when evaluating using the AUROC, AUPR and excess-AURC metrics, and report the corresponding results in the supplementary.

\begin{table}[t]
    \centering
    \input{tab_methods1_original_post.tex}
    \vspace{-10pt}
    \caption{\textbf{Confidence estimation performance on the SVHN, CIFAR-10 \& CIFAR-100 validation sets images.} Comparing AORC scores corresponding to the baseline method of Hendrcks \& Gimpel \yrcite{hendrycks2016msr} (MSR($x$)) with scores corresponding to our method with (BS($D_\chi$)) and without (MSR($D_\chi$)) using bootstrapping. Presented AORC values are multiplied by $100$ for clarity. Values in parenthesis are relative to the baseline performance.}
    \vspace{-25pt}
    \label{tab:methods_by_original}
\end{table}{}

Tab.~\ref{tab:methods_by_original} presents a comparison of our method with the baseline MSR (1\textsuperscript{st} row) on the SVHN, CIFAR-10 and CIFAR-100 datasets of small ($32\times32$ pixels) images, demonstrating the advantage of our method (with and without bootstrapping) over the baseline in all of them. A comparison on 49,000 ImageNet images\footnote{The remaining 1000 images of the ImageNet validation set were used as our experimental subset.} is brought in Tab.~\ref{tab:methods_by_original_ImageNet} for the top-1 and top-5 classification tasks, using the top performing ResNext classifier, as well as the less accurate ResNet18 and AlexNet architectures (classifier accuracy values are presented in the table). AORC values indicate that using our approach (MSR($D_\chi$)) consistently yields significant performance gains, across a wide range of datasets and classifier's qualities. Using the bootstrapping variant of our method (BS($D_\chi$)) further improves performance in almost all cases.

\begin{table*}[t]
    \centering
    \input{tab_methods2_original_post.tex}
    \vspace{-5pt}
    \caption{\textbf{Performance on the ImageNet validation set images.} We use the best, worst and intermediate performing pre-trained image classifiers available by PyTorch\footnote{https://pytorch.org/docs/stable/torchvision/models.html}. Performance is evaluated on the top-1 and top-5 classification tasks, after adapting the compared methods and the AORC score for the top-5 case (Please see details in supplementary). Please refer to the caption of Tab.~\ref{tab:methods_by_original} for more details.}
    \vspace{-5pt}
    \label{tab:methods_by_original_ImageNet}
\end{table*}{}

The results of our comparison on the STL-10 dataset of natural images ($96\times96$ pixels) are reported in Tab.~\ref{tab:methods_by_original_STL10}. We begin by experimenting on an ELU architecture classifier trained by Mandlebaum \& Weinshall, to allow comparison with their internal activation distances (IAD) based confidence estimation method \yrcite{mandelbaum2017distance_confidence}, that unlike ours requires adapting the classifier to allow accessing its internal layers. Note that while their method outperforms the baseline (possibly due to its access to internal classifier activations), our method does better still\footnote{To allow a fair comparison, we used the model from \cite{mandelbaum2017distance_confidence} that was trained using the conventional cross-entropy loss, and not the one that was specifically trained to optimize the IAD method's performance.}.

\begin{table}[h]
    \centering
    \input{tab_methods_stl10_original_post.tex}
    \vspace{-5pt}
    \caption{\textbf{Confidence estimation performance on the STL-10 validation set images.} We compare the baseline MSR method \cite{hendrycks2016msr} (MSR($x$)) and the invasive method by Mandelbaum \& Weinshall \yrcite{mandelbaum2017distance_confidence} (IAD) with a variant of our method utilizing only the original and horizontally flipped versions of $x$ (MSR($x$,$\Leftrightarrow$)), as well as with our method with and without bootstrapping (last two rows). Please refer to the text for a discussion and details about the classifiers, and to the caption of Tab.~\ref{tab:methods_by_original} for details about the display format.}
    \vspace{-5pt}
    \label{tab:methods_by_original_STL10}
\end{table}{}

When evaluating on the STL-10 dataset, we take advantage of its relatively small size, allowing us to easily conduct some experiments that require training new classifiers. The two right most columns of Tab.~\ref{tab:methods_by_original_STL10} correspond to evaluations on the two Wide-ResNet models trained with vs. without horizontal flipping data-augmentation (both models employed random shifting data-augmentation of up to $12$ pixels). 
The MSR($x,\Leftrightarrow$) row corresponds to estimating confidence using the original image and its horizontally flipped version alone. Values in this row suggest that incorporating a transformation into the data augmentation procedure has some negative effect on the ability to estimate classification confidence using this transformation (despite outperforming the baseline even in this special case). 
The effect of the specific choice of transformations utilized during classifier's training decreases when estimation is based on the entire set of transformations ${\cal T}$ ($|{\cal T}|=33$ in this case), as indicated by the values in the bottom two rows.

\begin{table}[t!]
    \centering
    \input{tab_models_ensemble_original_post.tex}
    \caption{\textbf{Performance when employing an ensemble of $5$ classifiers on the STL-10 dataset.} While AORC scores corresponding to using the ensemble (left column) exceed those achieved using a single classifier (right column), the benefit of using the proposed method is evident in both cases. Please refer to the caption of Tab.~\ref{tab:methods_by_original} for more details.}
    \vspace{-15pt}
    \label{tab:methods_by_T_ens}
\end{table}{}

Finally, we examined the performance of our method in the special case of using an ensemble of classifiers, as proposed by \cite{lakshminarayanan2017deep_ensembles}. To this end, we followed the procedure of \cite{geifman2018biasreduced} and trained additional $4$ Wide-ResNet classifiers, using different seeds for their weights initialization. In accordance with previous findings, the first row of Tab.~\ref{tab:methods_by_T_ens} indicates the resulting ensemble of $5$ classifiers (left column) outperforms the baseline confidence estimation performance (right column). 
Nonetheless, the results indicate that combining the NN-ens method \cite{lakshminarayanan2017deep_ensembles} with ours (with or without bootstrapping), by averaging the 5 classifiers' outputs corresponding to the transformed image set $D_\chi$, yields an additional performance gain.

To demonstrate the significance of our method's advantage, let us examine the RC curve in Fig.~\ref{fig:RC_example}, simulating a selective classification task over 22778 SVHN validation set images of house number digits, classified using a pre-trained VGG-8 with $0.896$ accuracy. choosing, e.g., a 1\% desired risk (horizontal dashed line) as our reliability threshold, using our method increases coverage from $0.17$ to $0.56$, which translates to classifying $9013$ more images with $99\%$ accuracy, compared to when using the original MSR method. Alternatively, we can pick a desired $0.56$ coverage value (vertical dashed line), in which case using our method leads to a $2.49$ points risk drop, corresponding to $567$ more correctly classified images. Even on datasets exhibiting smaller performance gaps, using our approach is still beneficial compared to the baseline. For instance, using the same $1\%$ desired risk threshold for classifying $49,000$ ImageNet images using the ResNext classifier yields a $1.7$ points coverage gain, which translates to classifying $818$ more images with $99\%$ accuracy.

Before concluding, we note that an additional avenue for further improvement is \emph{learning} a confidence estimator, beyond simple statistical estimates developed here. Our preliminary experiments on this are promising.

\input{fig_RC_example.tex}

%% file: fig_AORC_demo.tex
\begin{figure}[b!]
\vspace{-10pt}
\centering
\hspace{-10pt}
\includegraphics[width=\columnwidth]{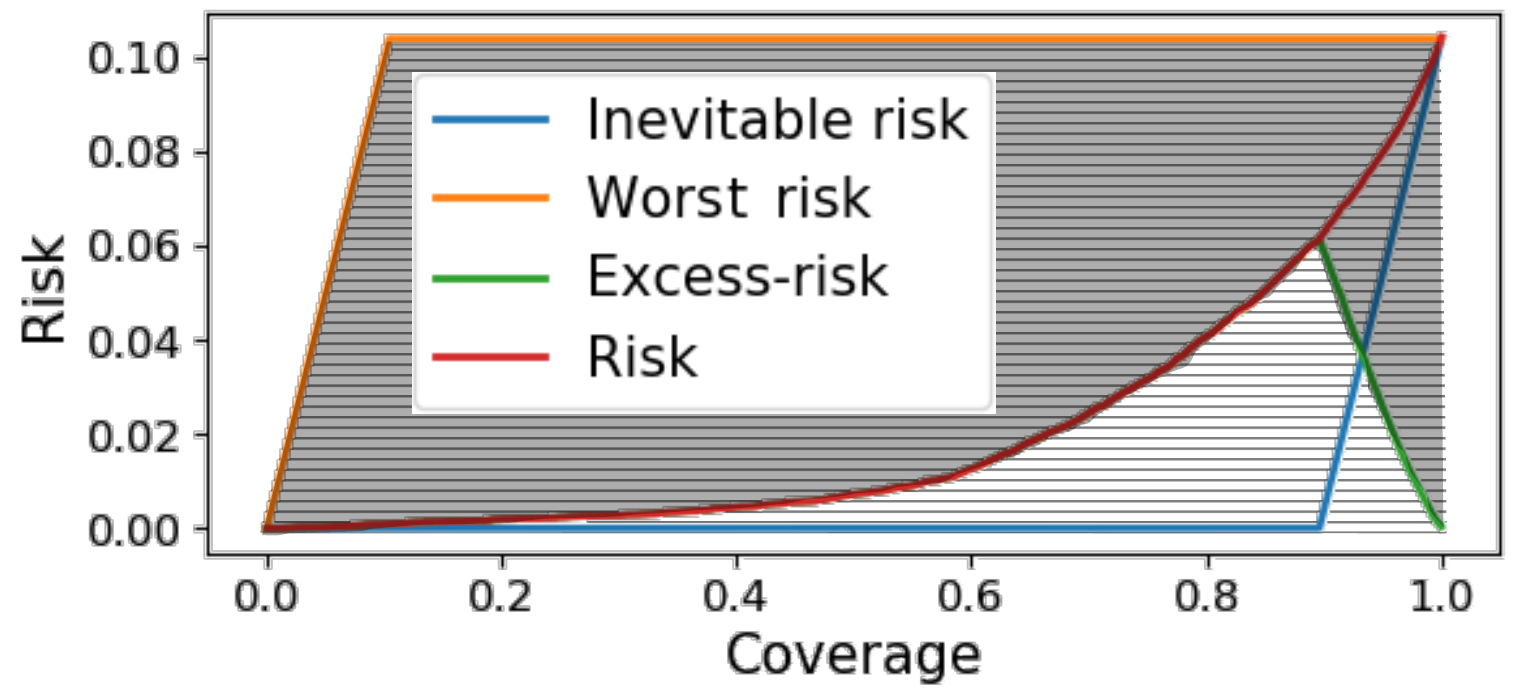}
\vspace{-10pt}
\caption{
\textbf{Explaining the area over risk-coverage (AORC) metric.} In a selective classification scenario (explained in Sec.~\ref{sec:selective_classification}), we subtract the inevitable cumulative risk incurred when ranking according to a faultless confidence estimator (blue) from the risk incurred using the evaluated confidence estimation method (red), yielding the excess risk curve (green). 
The AORC is then calculated by dividing the dark striped area, encapsulated between the worst (orange) and excess risk curves, by the entire striped area residing below the worst risk curve. An ideal confidence estimation method would yield the blue curve, resulting in an AORC=$1$, while the worst possible method would yield the orange curve with its AORC=$0$ score. (The classifier's accuracy in this illustration is slightly below $0.9$.)}

\label{fig:AORC_demo}
\end{figure}

%% file: fig_per_param_AORC.tex
\begin{figure}[ht!]
\centering
\hspace{-10pt}{
\includegraphics[width=\columnwidth]{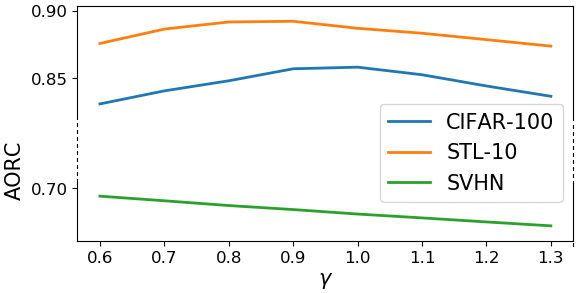}
\includegraphics[width=0.98\columnwidth]{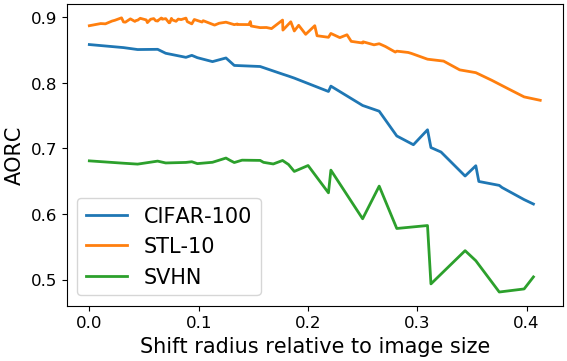}}
\vspace{-5pt}
\caption{
\textbf{AORC vs. transformation intensity.} 
Confidence estimation performance corresponding to different $\gamma$ (top plot) or image shifting magnitude (bottom plot) values, when estimating by applying MSR on a single Gamma transformed or shifting transformed image version, respectively. For the shift transformation (bottom), per-magnitude values are averaged over different shifting directions. For the CIFAR-100 and STL-10 datasets, each experiment is repeated with a preliminary horizontal shift, and AORC values corresponding to the two versions are averaged.
\vspace{-15pt}
\label{fig:per_param_AORC}}
\end{figure}

%% file: tab_transformations_original_post.tex
\begin{tabular}{lcccc}
\toprule
Dataset &  SVHN & CIFAR-100 & \multicolumn{2}{c}{STL-10} \\
$\Longleftrightarrow$~D.A. &     ~ &         ~ &     w/ &   w/o \\
Accuracy & 0.905 &     0.727 &  0.888 & 0.864 \\
\midrule
$x$                                           & 68.10 &     85.83 &  88.70 & 89.93 \\
$\Longleftrightarrow$                         & 38.55 &     85.31 &  89.02 & 86.75 \\
$\downarrow5$                                 & 67.99 &     82.07 &  88.71 & 90.84 \\
$\uparrow5$                                   & 67.36 &     81.01 &  89.63 & 90.40 \\
$\Longleftrightarrow$+$\uparrow5$             & 40.71 &     81.83 &  89.80 & 87.36 \\
$3^\circ\circlearrowright$                    & 66.64 &     80.42 &  81.68 & 83.24 \\
$\Updownarrow$                                & 39.37 &     55.45 &  56.27 & 60.78 \\
{\color{red}B}{\color{green}G}{\color{blue}R} & 67.29 &     63.43 &  77.85 & 82.22 \\
\bottomrule
\end{tabular}

%% file: tab_methods1_original_post.tex
\begin{tabular}{lccc}
\toprule
Dataset &              CIFAR-10 &             CIFAR-100 &                   SVHN \\
Arch. &                 VGG-8 &                 VGG-8 &                  VGG-8 \\
Accuracy &                 0.938 &                 0.744 &                  0.896 \\
\midrule
MSR($x$)      &          93.0 (+0.00) &          86.8 (+0.00) &           69.0 (+0.00) \\
MSR($D_\chi$) &          94.1 (+1.10) &          88.1 (+1.36) & 83.9 \textbf{(+14.85)} \\
BS($D_\chi$)  & 94.2 \textbf{(+1.24)} & 88.9 \textbf{(+2.09)} &          81.2 (+12.15) \\
\bottomrule
\end{tabular}

%% file: tab_methods2_original_post.tex
\begin{tabular}{lcccccc}
\toprule
Task & \multicolumn{3}{c}{Top-1} & \multicolumn{3}{c}{Top-5} \\
Arch. &                AlexNet &               ResNet18 &                ResNext &                AlexNet &               ResNet18 &                ResNext \\
Accuracy &                  0.565 &                  0.697 &                  0.793 &                  0.790 &                  0.891 &                  0.945 \\
\midrule
MSR($x$)      &          84.89 (+0.00) &          85.76 (+0.00) &          86.23 (+0.00) &          85.13 (+0.00) &          86.79 (+0.00) &          87.50 (+0.00) \\
MSR($D_\chi$) &          86.27 (+1.38) &          86.92 (+1.16) &          87.31 (+1.08) &          85.72 (+0.60) &          87.34 (+0.55) &          88.49 (+0.99) \\
BS($D_\chi$)  & 86.36 \textbf{(+1.47)} & 87.05 \textbf{(+1.28)} & 87.35 \textbf{(+1.12)} & 85.85 \textbf{(+0.72)} & 87.45 \textbf{(+0.67)} & 88.51 \textbf{(+1.01)} \\
\bottomrule
\end{tabular}

%% file: tab_methods_stl10_original_post.tex
\begin{tabular}{lccc}
\toprule
Arch. &                   ELU &           \multicolumn{2}{c}{Wide-ResNet} \\
$\Leftrightarrow$~D.A. &                   ~w/ &                    w/ &                   w/o \\
Accuracy &                 0.684 &                 0.891 &                 0.878 \\
\midrule
MSR($x$)                   &          77.9 (+0.00) &          91.7 (+0.00) &          91.3 (+0.00) \\
IAD                        &          78.2 (+0.31) &                       &                       \\
MSR($x$,$\Leftrightarrow$) &          79.3 (+1.41) &          92.0 (+0.25) &          92.7 (+1.39) \\
MSR($D_\chi$)              &          81.3 (+3.37) &          93.0 (+1.28) &          93.1 (+1.84) \\
BS($D_\chi$)               & 81.9 \textbf{(+4.04)} & 93.0 \textbf{(+1.33)} & 93.2 \textbf{(+1.88)} \\
\bottomrule
\end{tabular}

%% file: tab_models_ensemble_original_post.tex
\begin{tabular}{lcc}
\toprule
Arch. &                NN-Ens &     Single Classifier \\
Accuracy &                 0.907 &                 0.891 \\
\midrule
MSR($x$)      &          92.2 (+0.00) &          91.7 (+0.00) \\
MSR($D_\chi$) &          93.2 (+0.92) &          93.0 (+1.28) \\
BS($D_\chi$)  & 93.2 \textbf{(+0.94)} & 93.0 \textbf{(+1.33)} \\
\bottomrule
\end{tabular}

%% file: fig_RC_example.tex
\begin{figure}[t!]
\centering
\hspace{-10pt}
\includegraphics[width=\columnwidth]{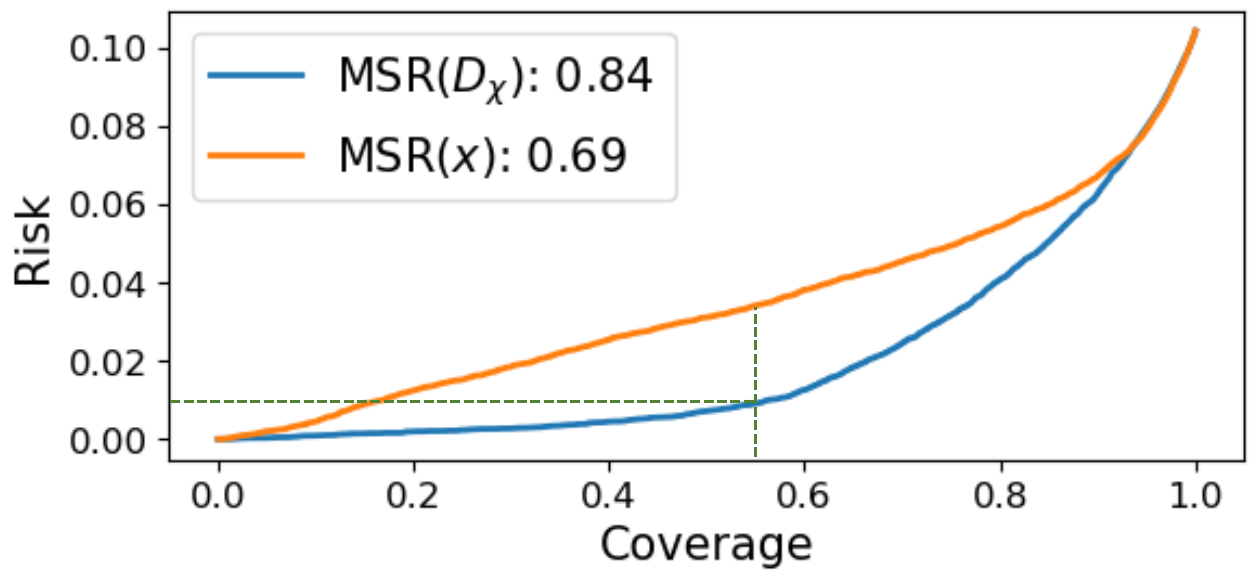}
\vspace{-10pt}
\caption{
\textbf{Selective classification example.} 
RC curves corresponding to selectively classifying 22778 SVHN validation set images, using a pre-trained VGG-8 classifier with $0.896$ accuracy, and ranking images' confidence estimated using our method (blue) vs the original MSR method (orange). Using our method yields a $0.39$ coverage gain ($9013$ images) for a 1\% desired risk choice (horizontal dashed line), or a $2.49$ risk drop ($567$ more correctly classified images) for the same coverage choice (vertical dashed line). Corresponding AORC values presented in the legend.}
\vspace{-10pt}
\label{fig:RC_example}
\end{figure}

%% file: conclusion.tex
\section{Conclusion}
 We present a novel, simple approach for estimating classification confidence on images. It can be applied to any trained classifier, without requiring modifications or access to its internal components. Our confidence estimation method is based on estimating the classifier's confidence on a set of images simulating capturing the original image content under different imaging conditions. These are synthesized using a family of image transformations commonly used for data augmentation. We evaluate the performance of our approach in the context of selective classification on a variety of datasets and classifiers, including ImageNet and ResNext, as well as ensembles of classifiers, demonstrating a consistent advantage over the previous state of the art in confidence estimation.

  We emphasize that the gains obtained by our method in this task do not come at a high computational expense, requiring only a modest overhead at test time, and no additional training. Combined with the importance of confidence estimation, in particular as a means to enable robust selective classification mechanisms with a reject option, this makes potential impact of our work and its extension significant.

%% file: supplementary_sections.tex
\appendix
\input{eval_params}
\input{performance_evaluation}
\input{top5_adaptation}
\section{Code and Datasets Partition}
The code for running our experiments on selected datasets and classifiers is available at \url{https://anonymous.4open.science/r/6cc90be7-d565-44c0-9397-182b196bd170/}, including files defining the partition into experimental and evaluation subsets for each dataset.

%% file: eval_params.tex
\section{Evaluation Parameters}
We next report the parameters $\cal T$, $N_{BS}$ and $W_{BS}$ used for the experimental evaluation presented in Tabs.~2-5 of the paper and in the coming Tabs.~\ref{tab:methods1_all_metrics}-\ref{tab:ensemble_all_metrics_D}. 

\subsection{Image Transformations} For each of the datasets we experimented with, we used the experimental images subset (see Sec.~4.2) to choose a set $\cal T$ yielding good confidence estimation performance, across the different applied classifiers. In the STL-10 dataset case, we used a separate transformation set for evaluating the ELU classifier (trained by Mandelbaum \& Weinshall \yrcite{mandelbaum2017distance_confidence}), as it operates on a downscaled version of the images ($32\times 32$ pixels) and not on the original ones (of size $96\times 96$ pixels).

We next report the transformation sets used for the experiments with each dataset. A `+' sign denotes consecutively applying image transformations, yielding a single image version, while different image versions (yielding different $x_i$s) are separated by a comma (`,'). We use the following notations to denote different transformation types: A horizontal image flip is denoted by a bidirectional arrow ($\Leftrightarrow$), and an image translation is denoted by a directional arrow followed by the translation magnitude (in pixels). A clockwise/anti-clockwise image rotation is denoted by the appropriate arrow ($\circlearrowright$~/~$\circlearrowleft$) followed by the rotation angle (in degrees), a zoom-in transformation is marked using the `$\times$' symbol followed by the magnification factor, and a gamma correction is denoted by a `$\gamma$' followed by the used gamma value.

\textbf{CIFAR-10} (Tabs.~2,~\ref{tab:methods1_all_metrics}~and~\ref{tab:methods1_all_metrics_D}):

($|{\cal T}|=13$)
$\cal T$~=~\{$\Leftrightarrow$+$\leftarrow 2$+$\downarrow 1$,~$\Leftrightarrow$+$\downarrow 1$,~$\Leftrightarrow$+$\uparrow 1$,~$\Leftrightarrow$+$\rightarrow 1$+$\downarrow 2$,~$\Leftrightarrow$+$\rightarrow 2$,~$\leftarrow 1$,~$\leftarrow 3$,~$\leftarrow 3$+$\uparrow 1$,~$\downarrow 1$,~$\rightarrow 1$+$\downarrow 1$,~$\rightarrow 1$,~$\rightarrow 1$+$\uparrow 1$\}

\textbf{CIFAR-100} (Tabs.~2,~\ref{tab:methods1_all_metrics}~and~\ref{tab:methods1_all_metrics_D}):

($|{\cal T}|=40$)
$\cal T$~=~\{$\Leftrightarrow$,~$\Leftrightarrow$+$\leftarrow 3$,~$\Leftrightarrow$+$\leftarrow 5$+$\downarrow 3$,~$\Leftrightarrow$+$\downarrow 2$,~$\Leftrightarrow$+$\downarrow 3$,~$\Leftrightarrow$+$\downarrow 5$,~$\Leftrightarrow$+$\uparrow 2$,~$\Leftrightarrow$+$\uparrow 5$,~$\Leftrightarrow$+$\rightarrow 11$+$\downarrow 11$,~$\Leftrightarrow$+$\rightarrow 12$+$\uparrow 12$,~$\Leftrightarrow$+$\rightarrow 14$+$\uparrow 14$,~$\Leftrightarrow$+$\rightarrow 1$+$\downarrow 3$,~$\Leftrightarrow$+$\rightarrow 1$,~$\Leftrightarrow$+$\rightarrow 2$+$\downarrow 3$,~$\Leftrightarrow$+$\rightarrow 2$,~$\Leftrightarrow$+$\rightarrow 2$+$\uparrow 1$,~$\Leftrightarrow$+$\rightarrow 2$+$\uparrow 2$,~$\Leftrightarrow$+$\rightarrow 4$,~$\Leftrightarrow$+$\rightarrow 4$+$\uparrow 4$,~$\Leftrightarrow$+$\rightarrow 5$+$\downarrow 11$,~$\Leftrightarrow$+$\rightarrow 5$+$\downarrow 5$,~$\Leftrightarrow$+$\rightarrow 6$,~$\Leftrightarrow$+$\rightarrow 7$,~$\circlearrowright 3$,~$\leftarrow 11$+$\downarrow 9$,~$\leftarrow 1$+$\downarrow 1$,~$\leftarrow 1$+$\downarrow 2$,~$\leftarrow 1$,~$\downarrow 1$,~$\downarrow 2$,~$\downarrow 3$,~$\uparrow 1$,~$\uparrow 5$,~$\rightarrow 10$+$\uparrow 10$,~$\rightarrow 14$+$\uparrow 14$,~$\rightarrow 1$+$\uparrow 2$,~$\rightarrow 1$+$\uparrow 3$,~$\rightarrow 2$+$\uparrow 1$,~$\rightarrow 9$+$\uparrow 11$\}

\textbf{SVHN} (Tabs.~2,~\ref{tab:methods1_all_metrics}~and~\ref{tab:methods1_all_metrics_D}):

($|{\cal T}|=17$)
$\cal T$~=~\{$\leftarrow 3$+$\downarrow 5$,~$\leftarrow 5$+$\downarrow 3$,~$\rightarrow 10$,~$\rightarrow 10$+$\uparrow 10$,~$\rightarrow 11$,~$\rightarrow 12$+$\uparrow 12$,~$\rightarrow 13$+$\uparrow 13$,~$\rightarrow 14$+$\uparrow 14$,~$\rightarrow 4$,~$\rightarrow 5$,~$\rightarrow 5$+$\uparrow 2$,~$\rightarrow 5$+$\uparrow 4$,~$\rightarrow 8$,~$\rightarrow 8$+$\uparrow 8$,~$\rightarrow 9$,~$\rightarrow 9$+$\uparrow 9$\}

\textbf{ImageNet} (Tabs.~3,~\ref{tab:methods2_all_metrics}~and~\ref{tab:methods2_all_metrics_D}):

($|{\cal T}|=24$)
$\cal T$~=~\{$\gamma 0.6$,~$\gamma 0.8$,~$\gamma 0.9$,~$\Leftrightarrow$+$\circlearrowleft 10$,~$\Leftrightarrow$+$\leftarrow 1$+$\uparrow 2$,~$\Leftrightarrow$+$\leftarrow 5$+$\uparrow 1$,~$\Leftrightarrow$+$\downarrow 10$,~$\Leftrightarrow$+$\uparrow 1$,~$\Leftrightarrow$+$\uparrow 10$,~$\Leftrightarrow$+$\uparrow 5$,~$\Leftrightarrow$+$\rightarrow 1$,~$\Leftrightarrow$+$\rightarrow 2$,~$\Leftrightarrow$+$\times 1.1$,~$\circlearrowright 1$,~$\circlearrowright 4$,~$\circlearrowright 5$,~$\circlearrowright 7$,~$\leftarrow 3$,~$\downarrow 10$,~$\uparrow 1$,~$\uparrow 2$,~$\times 1.1$,~$\times 1.1$+$\gamma 0.8$\}

\textbf{STL-10 (Wide-ResNet, including ensemble of classifiers)} (Tabs.~4,~5,~\ref{tab:stl10_all_metrics},~\ref{tab:ensemble_all_metrics},~\ref{tab:stl10_all_metrics_D}~and~\ref{tab:ensemble_all_metrics_D}):

($|{\cal T}|=35$)
$\cal T$~=~\{$\gamma 0.6$,~$\gamma 0.8$,~$\Leftrightarrow$+$\gamma 0.6$,~$\Leftrightarrow$+$\circlearrowleft 1$,~$\Leftrightarrow$+$\circlearrowleft 7$,~$\Leftrightarrow$+$\circlearrowright 1$,~$\Leftrightarrow$+$\leftarrow 3$+$\uparrow 6$,~$\Leftrightarrow$+$\leftarrow 4$+$\downarrow 2$,~$\Leftrightarrow$+$\leftarrow 4$+$\uparrow 6$,~$\Leftrightarrow$+$\uparrow 12$,~$\Leftrightarrow$+$\rightarrow 12$,~$\Leftrightarrow$+$\rightarrow 12$+$\uparrow 12$,~$\Leftrightarrow$+$\rightarrow 13$,~$\Leftrightarrow$+$\rightarrow 14$,~$\Leftrightarrow$+$\rightarrow 19$,~$\Leftrightarrow$+$\rightarrow 4$+$\uparrow 1$,~$\Leftrightarrow$+$\rightarrow 4$+$\uparrow 6$,~$\Leftrightarrow$+$\rightarrow 5$+$\downarrow 2$,~$\Leftrightarrow$+$\rightarrow 5$+$\uparrow 6$,~$\Leftrightarrow$+$\rightarrow 6$+$\downarrow 3$,~$\Leftrightarrow$+$\rightarrow 7$+$\downarrow 4$,~$\Leftrightarrow$+$\rightarrow 7$,~$\circlearrowright 1$,~$\leftarrow 4$+$\uparrow 6$,~$\leftarrow 6$+$\uparrow 6$,~$\leftarrow 9$+$\uparrow 9$,~$\uparrow 10$,~$\uparrow 12$,~$\uparrow 14$,~$\uparrow 18$,~$\uparrow 22$,~$\rightarrow 14$+$\uparrow 14$,~$\rightarrow 22$+$\uparrow 22$\}

\textbf{STL-10 (ELU)} (Tabs.~4,~\ref{tab:stl10_all_metrics}~and~\ref{tab:stl10_all_metrics_D}):

($|{\cal T}|=13$)
$\cal T$~=~\{$\gamma 1.2$,~$\Leftrightarrow$+$\downarrow 1$,~$\Leftrightarrow$+$\downarrow 5$,~$\Leftrightarrow$+$\rightarrow 1$+$\uparrow 1$,~$\Leftrightarrow$+$\times 1.1$,~$\circlearrowright 5$,~$\downarrow 1$,~$\downarrow 5$,~$\downarrow 7$,~$\rightarrow 1$,~$\times 1.1$,~$\times 1.1$+$\uparrow 5$\}

\subsection{Bootstrapping Parameters}
When evaluating our bootstrapping configuration (denoted BS($D_\chi$)), we set the number of resamplings from $\cal T$ (see Sec.~3.1) to \mbox{$N_{BS}=\min\{1000,\max\{100,0.001\cdot g(|\cal T|)\}\}$}, where $g(|\cal T|)$ is the number of possible different resampling options, calculated as \mbox{$g(|{\cal T}|)=\frac{(2|{\cal T}|-1)!}{(|{\cal T}|)!\cdot(|{\cal T}|-1)!}$}. The plurality window length in our sliding plurality window algorithm (Sec.~3.2) was empirically set to $W_{BS}=N_{BS}$ in all experiments. As with the chosen transformation sets, these parameters were empirically chosen so as to yield good confidence estimation performance (using our method's bootstrapping configuration) on the experimental images subsets.

%% file: performance_evaluation.tex
\section{Alternative Evaluation Metrics}
For evaluating the performance of our confidence estimation method, we introduce in Sec.~4.1 the AORC metric, a slightly modified version of the recently introduced excess Area Under Risk-Coverage curve (eAURC) metric \cite{geifman2018biasreduced}. However, to show that the advantage of our method remains unchanged when evaluating using the eAURC metric and the traditional Area Under Precision-Recall curve (AUPR) and Area Under Receiver Operator Characteristics curve (AUROC) metrics, we next present tables comparing performance using each of these 4 metrics (including AORC, which was already reported in the paper). Note that in all metrics but eAURC, a higher score corresponds to better confidence estimation performance. When using eAURC, lower scores correspond to better performance.

\begin{table}[hbt!]
    \centering
    \subfloat[\textbf{AORC.} These scores are already presented in Tab.~\ref{tab:methods_by_original} in the paper.]{
    \input{tables/AORC_tab_methods1_original_post}}
    \subfloat[\textbf{eAURC scores.} Here lower is better.]{
    \input{tables/eAURC_tab_methods1_original_post}}\\
    \subfloat[\textbf{AUPR}]{
    \input{tables/AUPR_tab_methods1_original_post}}
    \subfloat[\textbf{AUROC}]{
    \input{tables/AUROC_tab_methods1_original_post}}
    \caption{\textbf{Performance on the CIFAR-10, CIFAR-100 and SVHN datasets.} Class prediction according to $x$. Please refer to Tab.~\ref{tab:methods_by_original} in the paper for more details.\label{tab:methods1_all_metrics}}
\end{table}{}
\begin{table}[hbt!]
    \centering
    \subfloat[\textbf{AORC.} These scores are already presented in Tab.~\ref{tab:methods_by_original_ImageNet} in the paper.]{
    \input{tables/AORC_tab_methods2_original_post}}\\
    \subfloat[\textbf{eAURC scores.} Here lower is better.]{
    \input{tables/eAURC_tab_methods2_original_post}}\\
    \subfloat[\textbf{AUPR}]{
    \input{tables/AUPR_tab_methods2_original_post}}\\
    \subfloat[\textbf{AUROC}]{
    \input{tables/AUROC_tab_methods2_original_post}}
    \caption{\textbf{Performance on the ImageNet validation set images, on the top-1 and top-5 classification tasks.} Class prediction according to $x$. Please refer to Tab.~\ref{tab:methods_by_original_ImageNet} in the paper for more details.\label{tab:methods2_all_metrics}}
\end{table}{}
\begin{table}[hbt!]
    \centering
    \subfloat[\textbf{AORC.} These scores are already presented in Tab.~\ref{tab:methods_by_original_STL10} in the paper.]{
    \input{tables/AORC_tab_methods_stl10_original_post}}
    \subfloat[\textbf{eAURC scores.} Here lower is better.]{
    \input{tables/eAURC_tab_methods_stl10_original_post}}\\
    \subfloat[\textbf{AUPR}]{
    \input{tables/AUPR_tab_methods_stl10_original_post}}
    \subfloat[\textbf{AUROC}]{
    \input{tables/AUROC_tab_methods_stl10_original_post}}
    \caption{\textbf{Performance on the STL-10 dataset.} Class prediction according to $x$. Please refer to Tab.~\ref{tab:methods_by_original_STL10} in the paper for more details.\label{tab:stl10_all_metrics}}
\end{table}{}
\begin{table}[hbt!]
    \centering
    \subfloat[\textbf{AORC.} These scores are already presented in Tab.~\ref{tab:methods_by_T_ens} in the paper.]{
    \input{tables/AORC_tab_models_ensemble_original_post}}
    \subfloat[\textbf{eAURC scores.} Here lower is better.]{
    \input{tables/eAURC_tab_models_ensemble_original_post}}\\
    \subfloat[\textbf{AUPR}]{
    \input{tables/AUPR_tab_models_ensemble_original_post}}
    \subfloat[\textbf{AUROC}]{
    \input{tables/AUROC_tab_models_ensemble_original_post}}
    \caption{\textbf{Performance on the STL-10 dataset using an ensemble of $5$ classifiers (left column), following the method by \cite{lakshminarayanan2017deep_ensembles}.} Class prediction according to $x$. Please refer to Tab.~\ref{tab:methods_by_T_ens} in the paper for more details.\label{tab:ensemble_all_metrics}}
\end{table}{}
\clearpage
\section{Performance When Classifying using all images in $D_\chi$}
As mentioned in Sec.~3, the transformed image versions $x_i\in D_\chi$ can be used in a test-time data augmentation scheme, in which the predicted class is chosen as \mbox{$\hat{c}_\chi(x) = \argmax_c\sum_{x_i\in D_\chi}s_c(x)$}. In this case, our confidence measure is modified to be \mbox{$\hat{r}_x^f=\frac{1}{|D_\chi|}\sum_{x_i\in D_\chi}s_{\hat{c}_\chi(x)}(x_i)$}, in order to reflect the estimated confidence of this different classification scheme.

We next repeat the comparison presented in Tabs.~\ref{tab:methods1_all_metrics}-\ref{tab:ensemble_all_metrics} for the case of classifying by all $x_i\in D_\chi$, to show that the advantage of our confidence estimation method is independent of the chosen classification scheme.

\begin{table}[h]
    \centering
    \subfloat[\textbf{AORC}]{
    \input{tables/AORC_tab_methods1_T_ensemble_post}}
    \subfloat[\textbf{eAURC scores.} Here lower is better.]{
    \input{tables/eAURC_tab_methods1_T_ensemble_post}}\\
    \subfloat[\textbf{AUPR}]{
    \input{tables/AUPR_tab_methods1_T_ensemble_post}}
    \subfloat[\textbf{AUROC}]{
    \input{tables/AUROC_tab_methods1_T_ensemble_post}}
    \caption{\textbf{Performance on the CIFAR-10, CIFAR-100 and SVHN datasets.} Class prediction based on all images $x_i\in D_\chi$. Please refer to Tab.~\ref{tab:methods_by_original} in the paper for more details.\label{tab:methods1_all_metrics_D}}
\end{table}{}
\begin{table}[h]
    \centering
    \subfloat[\textbf{AORC}]{
    \input{tables/AORC_tab_methods2_T_ensemble_post}}\\
    \subfloat[\textbf{eAURC scores.} Here lower is better.]{
    \input{tables/eAURC_tab_methods2_T_ensemble_post}}\\
    \subfloat[\textbf{AUPR}]{
    \input{tables/AUPR_tab_methods2_T_ensemble_post}}\\
    \subfloat[\textbf{AUROC}]{
    \input{tables/AUROC_tab_methods2_T_ensemble_post}}
    \caption{\textbf{Performance on the ImageNet validation set images, on the top-1 and top-5 classification tasks.} Class prediction based on all images $x_i\in D_\chi$. Please refer to Tab.~\ref{tab:methods_by_original_ImageNet} in the paper for more details.\label{tab:methods2_all_metrics_D}}
\end{table}{}
\begin{table}[h]
    \centering
    \subfloat[\textbf{AORC}]{
    \input{tables/AORC_tab_methods_stl10_T_ensemble_post}}
    \subfloat[\textbf{eAURC scores.} Here lower is better.]{
    \input{tables/eAURC_tab_methods_stl10_T_ensemble_post}}\\
    \subfloat[\textbf{AUPR}]{
    \input{tables/AUPR_tab_methods_stl10_T_ensemble_post}}
    \subfloat[\textbf{AUROC}]{
    \input{tables/AUROC_tab_methods_stl10_T_ensemble_post}}
    \caption{\textbf{Performance on the STL-10 dataset.} Class prediction based on all images $x_i\in D_\chi$. Please refer to Tab.~\ref{tab:methods_by_original_STL10} in the paper for more details.\label{tab:stl10_all_metrics_D}}
\end{table}{}
\begin{table}[h]
    \centering
    \subfloat[\textbf{AORC}]{
    \input{tables/AORC_tab_models_ensemble_T_ensemble_post}}
    \subfloat[\textbf{eAURC scores.} Here lower is better.]{
    \input{tables/eAURC_tab_models_ensemble_T_ensemble_post}}\\
    \subfloat[\textbf{AUPR}]{
    \input{tables/AUPR_tab_models_ensemble_T_ensemble_post}}
    \subfloat[\textbf{AUROC}]{
    \input{tables/AUROC_tab_models_ensemble_T_ensemble_post}}
    \caption{\textbf{Performance on the STL-10 dataset using an ensemble of $5$ classifiers (left column), following the method by \cite{lakshminarayanan2017deep_ensembles}.} Class prediction based on all images $x_i\in D_\chi$. Please refer to Tab.~\ref{tab:methods_by_T_ens} in the paper for more details.\label{tab:ensemble_all_metrics_D}}
\end{table}{}
\clearpage

%% file: tables/AORC_tab_methods1_original_post.tex
\begin{tabular}{lccc}
\toprule
Dataset &              CIFAR-10 &             CIFAR-100 &                   SVHN \\
Arch. &                 VGG-8 &                 VGG-8 &                  VGG-8 \\
Accuracy &                 0.938 &                 0.744 &                  0.896 \\
\midrule
MSR($x$)      &          93.0 (+0.00) &          86.8 (+0.00) &           69.0 (+0.00) \\
MSR($D_\chi$) &          94.1 (+1.10) &          88.1 (+1.36) & 83.9 \textbf{(+14.85)} \\
BS($D_\chi$)  & 94.2 \textbf{(+1.24)} & 88.9 \textbf{(+2.09)} &          81.2 (+12.15) \\
\bottomrule
\end{tabular}

%% file: tables/eAURC_tab_methods1_original_post.tex
\begin{tabular}{lccc}
\toprule
Dataset &             CIFAR-10 &            CIFAR-100 &                 SVHN \\
Arch. &                VGG-8 &                VGG-8 &                VGG-8 \\
Accuracy &                0.938 &                0.744 &                0.896 \\
\midrule
MSR($x$)      &          0.4 (+0.00) &          2.5 (+0.00) &          2.9 (+0.00) \\
MSR($D_\chi$) &          0.3 (-0.06) &          2.3 (-0.26) & 1.5 \textbf{(-1.39)} \\
BS($D_\chi$)  & 0.3 \textbf{(-0.07)} & 2.1 \textbf{(-0.40)} &          1.8 (-1.14) \\
\bottomrule
\end{tabular}

%% file: tables/AUPR_tab_methods1_original_post.tex
\begin{tabular}{lccc}
\toprule
Dataset &              CIFAR-10 &             CIFAR-100 &                  SVHN \\
Arch. &                 VGG-8 &                 VGG-8 &                 VGG-8 \\
Accuracy &                 0.938 &                 0.744 &                 0.896 \\
\midrule
MSR($x$)      &          41.7 (+0.00) &          64.1 (+0.00) &          31.5 (+0.00) \\
MSR($D_\chi$) & 50.2 \textbf{(+8.48)} &          72.7 (+8.65) & 38.6 \textbf{(+7.07)} \\
BS($D_\chi$)  &          50.2 (+8.47) & 73.1 \textbf{(+9.00)} &          36.4 (+4.89) \\
\bottomrule
\end{tabular}

%% file: tables/AUROC_tab_methods1_original_post.tex
\begin{tabular}{lccc}
\toprule
Dataset &              CIFAR-10 &             CIFAR-100 &                   SVHN \\
Arch. &                 VGG-8 &                 VGG-8 &                  VGG-8 \\
Accuracy &                 0.938 &                 0.744 &                  0.896 \\
\midrule
MSR($x$)      &          93.0 (+0.00) &          86.8 (+0.00) &           69.0 (+0.00) \\
MSR($D_\chi$) &          94.1 (+1.10) &          88.1 (+1.36) & 83.9 \textbf{(+14.85)} \\
BS($D_\chi$)  & 94.2 \textbf{(+1.24)} & 88.9 \textbf{(+2.09)} &          81.2 (+12.15) \\
\bottomrule
\end{tabular}

%% file: tables/AORC_tab_methods2_original_post.tex
\begin{tabular}{lcccccc}
\toprule
Task & \multicolumn{3}{c}{Top-1} & \multicolumn{3}{c}{Top-5} \\
Arch. &                AlexNet &               ResNet18 &                ResNext &                AlexNet &               ResNet18 &                ResNext \\
Accuracy &                  0.565 &                  0.697 &                  0.793 &                  0.790 &                  0.891 &                  0.945 \\
\midrule
MSR($x$)      &          84.89 (+0.00) &          85.76 (+0.00) &          86.23 (+0.00) &          85.13 (+0.00) &          86.79 (+0.00) &          87.50 (+0.00) \\
MSR($D_\chi$) &          86.27 (+1.38) &          86.92 (+1.16) &          87.31 (+1.08) &          85.72 (+0.60) &          87.34 (+0.55) &          88.49 (+0.99) \\
BS($D_\chi$)  & 86.36 \textbf{(+1.47)} & 87.05 \textbf{(+1.28)} & 87.35 \textbf{(+1.12)} & 85.85 \textbf{(+0.72)} & 87.45 \textbf{(+0.67)} & 88.51 \textbf{(+1.01)} \\
\bottomrule
\end{tabular}

%% file: tables/eAURC_tab_methods2_original_post.tex
\begin{tabular}{lcccccc}
\toprule
Task & \multicolumn{3}{c}{Top-1} & \multicolumn{3}{c}{Top-5} \\
Arch. &               AlexNet &              ResNet18 &               ResNext &               AlexNet &              ResNet18 &               ResNext \\
Accuracy &                 0.565 &                 0.697 &                 0.793 &                 0.790 &                 0.891 &                 0.945 \\
\midrule
MSR($x$)      &          3.71 (+0.00) &          3.01 (+0.00) &          2.26 (+0.00) &          2.47 (+0.00) &          1.29 (+0.00) &          0.65 (+0.00) \\
MSR($D_\chi$) &          3.38 (-0.34) &          2.76 (-0.24) &          2.09 (-0.18) &          2.37 (-0.10) &          1.23 (-0.05) &          0.60 (-0.05) \\
BS($D_\chi$)  & 3.35 \textbf{(-0.36)} & 2.74 \textbf{(-0.27)} & 2.08 \textbf{(-0.18)} & 2.35 \textbf{(-0.12)} & 1.22 \textbf{(-0.07)} & 0.59 \textbf{(-0.05)} \\
\bottomrule
\end{tabular}

%% file: tables/AUPR_tab_methods2_original_post.tex
\begin{tabular}{lcccccc}
\toprule
Task & \multicolumn{3}{c}{Top-1} & \multicolumn{3}{c}{Top-5} \\
Arch. &                AlexNet &               ResNet18 &                ResNext &                AlexNet &               ResNet18 &                ResNext \\
Accuracy &                  0.565 &                  0.697 &                  0.793 &                  0.790 &                  0.891 &                  0.945 \\
\midrule
MSR($x$)      &          78.69 (+0.00) &          70.18 (+0.00) &          61.17 (+0.00) &          57.37 (+0.00) &          45.56 (+0.00) &          34.22 (+0.00) \\
MSR($D_\chi$) &          81.31 (+2.63) &          73.57 (+3.39) &          64.82 (+3.65) &          59.51 (+2.14) &          47.98 (+2.42) &          37.84 (+3.62) \\
BS($D_\chi$)  & 81.34 \textbf{(+2.65)} & 73.64 \textbf{(+3.46)} & 64.92 \textbf{(+3.75)} & 59.63 \textbf{(+2.26)} & 48.11 \textbf{(+2.55)} & 37.87 \textbf{(+3.65)} \\
\bottomrule
\end{tabular}

%% file: tables/AUROC_tab_methods2_original_post.tex
\begin{tabular}{lcccccc}
\toprule
Task & \multicolumn{3}{c}{Top-1} & \multicolumn{3}{c}{Top-5} \\
Arch. &                AlexNet &               ResNet18 &                ResNext &                AlexNet &               ResNet18 &                ResNext \\
Accuracy &                  0.565 &                  0.697 &                  0.793 &                  0.790 &                  0.891 &                  0.945 \\
\midrule
MSR($x$)      &          84.89 (+0.00) &          85.76 (+0.00) &          86.23 (+0.00) &          85.13 (+0.00) &          86.79 (+0.00) &          87.50 (+0.00) \\
MSR($D_\chi$) &          86.27 (+1.38) &          86.92 (+1.16) &          87.31 (+1.08) &          85.72 (+0.60) &          87.34 (+0.55) &          88.49 (+0.99) \\
BS($D_\chi$)  & 86.36 \textbf{(+1.47)} & 87.05 \textbf{(+1.28)} & 87.35 \textbf{(+1.12)} & 85.85 \textbf{(+0.72)} & 87.45 \textbf{(+0.67)} & 88.51 \textbf{(+1.01)} \\
\bottomrule
\end{tabular}

%% file: tables/AORC_tab_methods_stl10_original_post.tex
\begin{tabular}{lccc}
\toprule
Arch. &                   ELU &           \multicolumn{2}{c}{Wide-ResNet} \\
$\Leftrightarrow$~D.A. &                   ~w/ &                    w/ &                   w/o \\
Accuracy &                 0.684 &                 0.891 &                 0.878 \\
\midrule
MSR($x$)      &          77.9 (+0.00) &          91.7 (+0.00) &          91.3 (+0.00) \\
IAD           &          78.2 (+0.31) &                       &                       \\
MSR($D_\chi$) &          81.3 (+3.37) &          93.0 (+1.28) &          93.1 (+1.84) \\
BS($D_\chi$)  & 81.9 \textbf{(+4.04)} & 93.0 \textbf{(+1.33)} & 93.2 \textbf{(+1.88)} \\
\bottomrule
\end{tabular}

%% file: tables/eAURC_tab_methods_stl10_original_post.tex
\begin{tabular}{lccc}
\toprule
Arch. &                  ELU &          \multicolumn{2}{c}{Wide-ResNet} \\
$\Leftrightarrow$~D.A. &                  ~w/ &                   w/ &                  w/o \\
Accuracy &                0.684 &                0.891 &                0.878 \\
\midrule
MSR($x$)      &          4.8 (+0.00) &          0.8 (+0.00) &          0.9 (+0.00) \\
IAD           &          4.7 (-0.07) &                      &                      \\
MSR($D_\chi$) &          4.1 (-0.73) &          0.7 (-0.12) &          0.7 (-0.20) \\
BS($D_\chi$)  & 3.9 \textbf{(-0.87)} & 0.7 \textbf{(-0.13)} & 0.7 \textbf{(-0.20)} \\
\bottomrule
\end{tabular}

%% file: tables/AUPR_tab_methods_stl10_original_post.tex
\begin{tabular}{lccc}
\toprule
Arch. &                   ELU &           \multicolumn{2}{c}{Wide-ResNet} \\
$\Leftrightarrow$~D.A. &                   ~w/ &                    w/ &                   w/o \\
Accuracy &                 0.684 &                 0.891 &                 0.878 \\
\midrule
MSR($x$)      &          56.7 (+0.00) &          53.6 (+0.00) &          56.3 (+0.00) \\
IAD           &          61.2 (+4.53) &                       &                       \\
MSR($D_\chi$) &          64.4 (+7.71) & 62.2 \textbf{(+8.55)} & 65.8 \textbf{(+9.50)} \\
BS($D_\chi$)  & 64.8 \textbf{(+8.13)} &          62.0 (+8.39) &          65.6 (+9.31) \\
\bottomrule
\end{tabular}

%% file: tables/AUROC_tab_methods_stl10_original_post.tex
\begin{tabular}{lccc}
\toprule
Arch. &                   ELU &           \multicolumn{2}{c}{Wide-ResNet} \\
$\Leftrightarrow$~D.A. &                   ~w/ &                    w/ &                   w/o \\
Accuracy &                 0.684 &                 0.891 &                 0.878 \\
\midrule
MSR($x$)      &          77.9 (+0.00) &          91.7 (+0.00) &          91.3 (+0.00) \\
IAD           &          78.2 (+0.31) &                       &                       \\
MSR($D_\chi$) &          81.3 (+3.37) &          93.0 (+1.28) &          93.1 (+1.84) \\
BS($D_\chi$)  & 81.9 \textbf{(+4.04)} & 93.0 \textbf{(+1.33)} & 93.2 \textbf{(+1.88)} \\
\bottomrule
\end{tabular}

%% file: tables/AORC_tab_models_ensemble_original_post.tex
\begin{tabular}{lcc}
\toprule
Arch. &                NN-Ens &     Single Classifier \\
Accuracy &                 0.907 &                 0.891 \\
\midrule
MSR($x$)      &          92.2 (+0.00) &          91.7 (+0.00) \\
MSR($D_\chi$) &          93.2 (+0.92) &          93.0 (+1.28) \\
BS($D_\chi$)  & 93.2 \textbf{(+0.94)} & 93.0 \textbf{(+1.33)} \\
\bottomrule
\end{tabular}

%% file: tables/eAURC_tab_models_ensemble_original_post.tex
\begin{tabular}{lcc}
\toprule
Arch. &               NN-Ens &    Single Classifier \\
Accuracy &                0.907 &                0.891 \\
\midrule
MSR($x$)      &          0.7 (+0.00) &          0.8 (+0.00) \\
MSR($D_\chi$) &          0.6 (-0.08) &          0.7 (-0.12) \\
BS($D_\chi$)  & 0.6 \textbf{(-0.08)} & 0.7 \textbf{(-0.13)} \\
\bottomrule
\end{tabular}

%% file: tables/AUPR_tab_models_ensemble_original_post.tex
\begin{tabular}{lcc}
\toprule
Arch. &                NN-Ens &     Single Classifier \\
Accuracy &                 0.907 &                 0.891 \\
\midrule
MSR($x$)      &          52.1 (+0.00) &          53.6 (+0.00) \\
MSR($D_\chi$) & 56.9 \textbf{(+4.75)} & 62.2 \textbf{(+8.55)} \\
BS($D_\chi$)  &          56.8 (+4.66) &          62.0 (+8.39) \\
\bottomrule
\end{tabular}

%% file: tables/AUROC_tab_models_ensemble_original_post.tex
\begin{tabular}{lcc}
\toprule
Arch. &                NN-Ens &     Single Classifier \\
Accuracy &                 0.907 &                 0.891 \\
\midrule
MSR($x$)      &          92.2 (+0.00) &          91.7 (+0.00) \\
MSR($D_\chi$) &          93.2 (+0.92) &          93.0 (+1.28) \\
BS($D_\chi$)  & 93.2 \textbf{(+0.94)} & 93.0 \textbf{(+1.33)} \\
\bottomrule
\end{tabular}

%% file: tables/AORC_tab_methods1_T_ensemble_post.tex
\begin{tabular}{lccc}
\toprule
Dataset &              CIFAR-10 &             CIFAR-100 &                  SVHN \\
Arch. &                 VGG-8 &                 VGG-8 &                 VGG-8 \\
Accuracy &                 0.940 &                 0.758 &                 0.766 \\
\midrule
MSR($x$)      &          92.8 (+0.00) &          85.1 (+0.00) &          70.2 (+0.00) \\
MSR($D_\chi$) &          93.8 (+0.98) &          86.8 (+1.66) & 77.0 \textbf{(+6.72)} \\
BS($D_\chi$)  & 93.9 \textbf{(+1.13)} & 87.6 \textbf{(+2.44)} &          73.9 (+3.64) \\
\bottomrule
\end{tabular}

%% file: tables/eAURC_tab_methods1_T_ensemble_post.tex
\begin{tabular}{lccc}
\toprule
Dataset &             CIFAR-10 &            CIFAR-100 &                 SVHN \\
Arch. &                VGG-8 &                VGG-8 &                VGG-8 \\
Accuracy &                0.940 &                0.758 &                0.766 \\
\midrule
MSR($x$)      &          0.4 (+0.00) &          2.7 (+0.00) &          5.3 (+0.00) \\
MSR($D_\chi$) &          0.4 (-0.06) &          2.4 (-0.30) & 4.1 \textbf{(-1.20)} \\
BS($D_\chi$)  & 0.3 \textbf{(-0.06)} & 2.3 \textbf{(-0.45)} &          4.7 (-0.65) \\
\bottomrule
\end{tabular}

%% file: tables/AUPR_tab_methods1_T_ensemble_post.tex
\begin{tabular}{lccc}
\toprule
Dataset &              CIFAR-10 &             CIFAR-100 &                  SVHN \\
Arch. &                 VGG-8 &                 VGG-8 &                 VGG-8 \\
Accuracy &                 0.940 &                 0.758 &                 0.766 \\
\midrule
MSR($x$)      &          39.6 (+0.00) &          58.1 (+0.00) &          49.9 (+0.00) \\
MSR($D_\chi$) &          48.7 (+9.10) &          67.6 (+9.43) &          52.8 (+2.84) \\
BS($D_\chi$)  & 48.7 \textbf{(+9.14)} & 68.1 \textbf{(+9.91)} & 53.1 \textbf{(+3.19)} \\
\bottomrule
\end{tabular}

%% file: tables/AUROC_tab_methods1_T_ensemble_post.tex
\begin{tabular}{lccc}
\toprule
Dataset &              CIFAR-10 &             CIFAR-100 &                  SVHN \\
Arch. &                 VGG-8 &                 VGG-8 &                 VGG-8 \\
Accuracy &                 0.940 &                 0.758 &                 0.766 \\
\midrule
MSR($x$)      &          92.8 (+0.00) &          85.1 (+0.00) &          70.2 (+0.00) \\
MSR($D_\chi$) &          93.8 (+0.98) &          86.8 (+1.66) & 77.0 \textbf{(+6.72)} \\
BS($D_\chi$)  & 93.9 \textbf{(+1.13)} & 87.6 \textbf{(+2.44)} &          73.9 (+3.64) \\
\bottomrule
\end{tabular}

%% file: tables/AORC_tab_methods2_T_ensemble_post.tex
\begin{tabular}{lcccccc}
\toprule
Task & \multicolumn{3}{c}{Top-1} & \multicolumn{3}{c}{Top-5} \\
Arch. &                AlexNet &               ResNet18 &                ResNext &                AlexNet &               ResNet18 &                ResNext \\
Accuracy &                  0.579 &                  0.708 &                  0.798 &                  0.801 &                  0.897 &                  0.949 \\
\midrule
MSR($x$)      &          83.59 (+0.00) &          84.69 (+0.00) &          85.60 (+0.00) &          84.57 (+0.00) &          85.93 (+0.00) &          86.92 (+0.00) \\
MSR($D_\chi$) &          84.82 (+1.23) &          85.74 (+1.05) &          86.64 (+1.04) &          85.12 (+0.55) &          86.44 (+0.52) &          87.76 (+0.84) \\
BS($D_\chi$)  & 84.94 \textbf{(+1.36)} & 85.89 \textbf{(+1.20)} & 86.69 \textbf{(+1.08)} & 85.26 \textbf{(+0.69)} & 86.57 \textbf{(+0.64)} & 87.80 \textbf{(+0.88)} \\
\bottomrule
\end{tabular}

%% file: tables/eAURC_tab_methods2_T_ensemble_post.tex
\begin{tabular}{lcccccc}
\toprule
Task & \multicolumn{3}{c}{Top-1} & \multicolumn{3}{c}{Top-5} \\
Arch. &               AlexNet &              ResNet18 &               ResNext &               AlexNet &              ResNet18 &               ResNext \\
Accuracy &                 0.579 &                 0.708 &                 0.798 &                 0.801 &                 0.897 &                 0.949 \\
\midrule
MSR($x$)      &          4.00 (+0.00) &          3.17 (+0.00) &          2.32 (+0.00) &          2.46 (+0.00) &          1.30 (+0.00) &          0.64 (+0.00) \\
MSR($D_\chi$) &          3.70 (-0.30) &          2.95 (-0.22) &          2.16 (-0.17) &          2.37 (-0.09) &          1.25 (-0.05) &          0.60 (-0.04) \\
BS($D_\chi$)  & 3.67 \textbf{(-0.33)} & 2.92 \textbf{(-0.25)} & 2.15 \textbf{(-0.17)} & 2.35 \textbf{(-0.11)} & 1.24 \textbf{(-0.06)} & 0.59 \textbf{(-0.04)} \\
\bottomrule
\end{tabular}

%% file: tables/AUPR_tab_methods2_T_ensemble_post.tex
\begin{tabular}{lcccccc}
\toprule
Task & \multicolumn{3}{c}{Top-1} & \multicolumn{3}{c}{Top-5} \\
Arch. &                AlexNet &               ResNet18 &                ResNext &                AlexNet &               ResNet18 &                ResNext \\
Accuracy &                  0.579 &                  0.708 &                  0.798 &                  0.801 &                  0.897 &                  0.949 \\
\midrule
MSR($x$)      &          75.50 (+0.00) &          66.31 (+0.00) &          58.08 (+0.00) &          54.45 (+0.00) &          41.47 (+0.00) &          30.74 (+0.00) \\
MSR($D_\chi$) &          77.77 (+2.27) &          69.32 (+3.01) &          61.41 (+3.32) &          56.41 (+1.96) &          43.40 (+1.93) &          33.28 (+2.54) \\
BS($D_\chi$)  & 77.91 \textbf{(+2.41)} & 69.47 \textbf{(+3.16)} & 61.47 \textbf{(+3.39)} & 56.65 \textbf{(+2.20)} & 43.63 \textbf{(+2.16)} & 33.35 \textbf{(+2.61)} \\
\bottomrule
\end{tabular}

%% file: tables/AUROC_tab_methods2_T_ensemble_post.tex
\begin{tabular}{lcccccc}
\toprule
Task & \multicolumn{3}{c}{Top-1} & \multicolumn{3}{c}{Top-5} \\
Arch. &                AlexNet &               ResNet18 &                ResNext &                AlexNet &               ResNet18 &                ResNext \\
Accuracy &                  0.579 &                  0.708 &                  0.798 &                  0.801 &                  0.897 &                  0.949 \\
\midrule
MSR($x$)      &          83.59 (+0.00) &          84.69 (+0.00) &          85.60 (+0.00) &          84.57 (+0.00) &          85.93 (+0.00) &          86.92 (+0.00) \\
MSR($D_\chi$) &          84.82 (+1.23) &          85.74 (+1.05) &          86.64 (+1.04) &          85.12 (+0.55) &          86.44 (+0.52) &          87.76 (+0.84) \\
BS($D_\chi$)  & 84.94 \textbf{(+1.36)} & 85.89 \textbf{(+1.20)} & 86.69 \textbf{(+1.08)} & 85.26 \textbf{(+0.69)} & 86.57 \textbf{(+0.64)} & 87.80 \textbf{(+0.88)} \\
\bottomrule
\end{tabular}

%% file: tables/AORC_tab_methods_stl10_T_ensemble_post.tex
\begin{tabular}{lccc}
\toprule
Arch. &                   ELU &\multicolumn{2}{c}{Wide-ResNet}\\
$\Leftrightarrow$~D.A. &                   ~w/ &                    w/ &                   w/o \\
Accuracy &                 0.703 &                 0.900 &                 0.890 \\
\midrule
MSR($x$)      &          75.8 (+0.00) &          90.6 (+0.00) &          89.4 (+0.00) \\
IAD           &          75.0 (-0.80) &                       &                       \\
MSR($D_\chi$) &          78.9 (+3.17) &          91.8 (+1.20) &          91.6 (+2.11) \\
BS($D_\chi$)  & 79.7 \textbf{(+3.95)} & 91.9 \textbf{(+1.26)} & 91.6 \textbf{(+2.11)} \\
\bottomrule
\end{tabular}

%% file: tables/eAURC_tab_methods_stl10_T_ensemble_post.tex
\begin{tabular}{lccc}
\toprule
Arch. &                  ELU &          \multicolumn{2}{c}{Wide-ResNet} \\
$\Leftrightarrow$~D.A. &                  ~w/ &                   w/ &                  w/o \\
Accuracy &                0.703 &                0.900 &                0.890 \\
\midrule
MSR($x$)      &          5.1 (+0.00) &          0.8 (+0.00) &          1.0 (+0.00) \\
IAD           &          5.2 (+0.17) &                      &                      \\
MSR($D_\chi$) &          4.4 (-0.66) &          0.7 (-0.11) &          0.8 (-0.21) \\
BS($D_\chi$)  & 4.2 \textbf{(-0.82)} & 0.7 \textbf{(-0.11)} & 0.8 \textbf{(-0.21)} \\
\bottomrule
\end{tabular}

%% file: tables/AUPR_tab_methods_stl10_T_ensemble_post.tex
\begin{tabular}{lccc}
\toprule
Arch. &                   ELU &           \multicolumn{2}{c}{Wide-ResNet} \\
$\Leftrightarrow$~D.A. &                   ~w/ &                    w/ &                   w/o \\
Accuracy &                 0.703 &                 0.900 &                 0.890 \\
\midrule
MSR($x$)      &          50.8 (+0.00) &          45.6 (+0.00) &          44.0 (+0.00) \\
IAD           &          50.7 (-0.16) &                       &                       \\
MSR($D_\chi$) &          57.3 (+6.51) & 51.7 \textbf{(+6.06)} & 52.2 \textbf{(+8.17)} \\
BS($D_\chi$)  & 58.1 \textbf{(+7.26)} &          51.6 (+5.97) &          51.9 (+7.89) \\
\bottomrule
\end{tabular}

%% file: tables/AUROC_tab_methods_stl10_T_ensemble_post.tex
\begin{tabular}{lccc}
\toprule
Arch. &                   ELU &          \multicolumn{2}{c}{Wide-ResNet} \\
$\Leftrightarrow$~D.A. &                   ~w/ &                    w/ &                   w/o \\
Accuracy &                 0.703 &                 0.900 &                 0.890 \\
\midrule
MSR($x$)      &          75.8 (+0.00) &          90.6 (+0.00) &          89.4 (+0.00) \\
IAD           &          75.0 (-0.80) &                       &                       \\
MSR($D_\chi$) &          78.9 (+3.17) &          91.8 (+1.20) &          91.6 (+2.11) \\
BS($D_\chi$)  & 79.7 \textbf{(+3.95)} & 91.9 \textbf{(+1.26)} & 91.6 \textbf{(+2.11)} \\
\bottomrule
\end{tabular}

%% file: tables/AORC_tab_models_ensemble_T_ensemble_post.tex
\begin{tabular}{lcc}
\toprule
Arch. &                NN-Ens &     Single Classifier \\
Accuracy &                 0.909 &                 0.900 \\
\midrule
MSR($x$)      &          91.9 (+0.00) &          90.6 (+0.00) \\
MSR($D_\chi$) &          92.8 (+0.89) &          91.8 (+1.20) \\
BS($D_\chi$)  & 92.8 \textbf{(+0.90)} & 91.9 \textbf{(+1.26)} \\
\bottomrule
\end{tabular}

%% file: tables/eAURC_tab_models_ensemble_T_ensemble_post.tex
\begin{tabular}{lcc}
\toprule
Arch. &               NN-Ens &    Single Classifier \\
Accuracy &                0.909 &                0.900 \\
\midrule
MSR($x$)      &          0.7 (+0.00) &          0.8 (+0.00) \\
MSR($D_\chi$) &          0.6 (-0.07) &          0.7 (-0.11) \\
BS($D_\chi$)  & 0.6 \textbf{(-0.07)} & 0.7 \textbf{(-0.11)} \\
\bottomrule
\end{tabular}

%% file: tables/AUPR_tab_models_ensemble_T_ensemble_post.tex
\begin{tabular}{lcc}
\toprule
Arch. &                NN-Ens &     Single Classifier \\
Accuracy &                 0.909 &                 0.900 \\
\midrule
MSR($x$)      &          49.7 (+0.00) &          45.6 (+0.00) \\
MSR($D_\chi$) & 53.7 \textbf{(+3.99)} & 51.7 \textbf{(+6.06)} \\
BS($D_\chi$)  &          53.6 (+3.88) &          51.6 (+5.97) \\
\bottomrule
\end{tabular}

%% file: tables/AUROC_tab_models_ensemble_T_ensemble_post.tex
\begin{tabular}{lcc}
\toprule
Arch. &                NN-Ens &     Single Classifier \\
Accuracy &                 0.909 &                 0.900 \\
\midrule
MSR($x$)      &          91.9 (+0.00) &          90.6 (+0.00) \\
MSR($D_\chi$) &          92.8 (+0.89) &          91.8 (+1.20) \\
BS($D_\chi$)  & 92.8 \textbf{(+0.90)} & 91.9 \textbf{(+1.26)} \\
\bottomrule
\end{tabular}

%% file: top5_adaptation.tex
\section{Adapting for the Top-5 Classification Task}
For evaluating our confidence estimation method on the top-5 ImageNet classification task (Tabs.~3,~\ref{tab:methods2_all_metrics}~and~\ref{tab:methods2_all_metrics_D}), we accommodated the notion of classification error and the MSR method to this particular case.
An image classification is considered to be erroneous in this case when neither of the leading 5 scores at the output of the classifier corresponds to the correct ground truth label of the image $c(x)$.

For confidence estimation, we modify the maximal softmax response (MSR) procedure by summing over softmax responses corresponding to the leading 5 scores, instead of taking only the maximal score. This modification applies both for the original MSR method by Hendrycks \& Gimpel \yrcite{hendrycks2016msr}, and for the computation of our classification confidence measure in Eq.~(2).